\pdfoutput=1

\documentclass[11pt]{article}

\usepackage{ACL2023}

\usepackage{times}
\usepackage{latexsym}

\usepackage[T1]{fontenc}

\usepackage[utf8]{inputenc}

\usepackage{microtype}

\usepackage{inconsolata}
\usepackage{cleveref}
\usepackage{graphicx}
\usepackage{booktabs}
\usepackage{subfigure}
\usepackage{multirow}
\usepackage{makecell}
\usepackage{enumitem}
\usepackage{float}

\usepackage{listings}
\usepackage{xcolor}

\definecolor{codegreen}{rgb}{0,0.6,0}
\definecolor{codegray}{rgb}{0.5,0.5,0.5}
\definecolor{codepurple}{rgb}{0.58,0,0.82}
\definecolor{backcolour}{rgb}{0.95,0.95,0.92}

\lstdefinestyle{mystyle}{
    backgroundcolor=\color{backcolour},   
    commentstyle=\color{codegreen},
    keywordstyle=\color{magenta},
    numberstyle=\tiny\color{codegray},
    stringstyle=\color{codepurple},
    basicstyle=\ttfamily\footnotesize,
    breakatwhitespace=false,         
    breaklines=true,                 
    captionpos=b,                    
    keepspaces=true,                 
    numbers=left,                    
    numbersep=5pt,                  
    showspaces=false,                
    showstringspaces=false,
    showtabs=false,                  
    tabsize=2
}

\title{{\tt SmartProbe}: A Virtual Moderator for Market Research Surveys}

\author{Josh Seltzer$^1$, Jiahua (Fiona) Pan$^1$, Kathy Cheng$^1$, Yuxiao Sun$^1$\\
\textbf{Santosh Kolagati$^1$, Jimmy Lin$^2$, Shi Zong$^2$}\\
$^1$Nexxt Intelligence\quad $^2$University of Waterloo\\
{\tt \{josh, fiona, kathy, shawn, santosh\}@nexxt.in} \\
{\tt \{jimmylin, s4zong\}@uwaterloo.ca}
}

\begin{document}

\maketitle

\begin{abstract}

Market research surveys are a powerful methodology for understanding consumer perspectives at scale, but are limited by depth of understanding and insights. A virtual moderator can introduce elements of qualitative research into surveys, developing a rapport with survey participants and dynamically asking probing questions, ultimately to elicit more useful information for market researchers. In this work, we introduce {\tt SmartProbe}, an API which leverages the adaptive capabilities of large language models (LLMs), and incorporates domain knowledge from market research, in order to generate effective probing questions in any market research survey. We outline the modular processing flow of {\tt SmartProbe}, and evaluate the quality and effectiveness of its generated probing questions. We believe our efforts will inspire industry practitioners to build real-world applications based on the latest advances in LLMs. Our demo is publicly available at \url{https://nexxt.in/smartprobe-demo}.
\end{abstract}

\section{Introduction}

Qualitative research, often conducted in the form of focus groups or in-depth interviews, is exploratory in nature and is used to understand the underlying reasons, opinions, and motivations behind a particular phenomenon. Quantitative research, predominantly conducted through online surveys, is used to measure and quantify baseline assumptions.

Market researchers have typically seen qualitative and quantitative research as two complimentary yet separate methodologies because of their fundamental differences in data collection and analysis. However, the need for combining qualitative depth and quantitative rigour in a hybrid research methodology has grown alongside increases in the speed of innovation and decision-making. With an emerging need for human-centered innovation, businesses are increasingly pressured to use consumer-led deeper insights at scale to fuel rapid iterations and developments.

Conducting mixed-methods research can be very time-consuming and resource-intensive. A technical breakthrough that combines both the advantages of qualitative and quantitative research is required, i.e., deeper qualitative insights at quantitative scale.

We propose that a ``virtual moderator'' can be used to bring qualitative insights to online surveys. Moderators (or interviewers) are trained market researchers who conduct group or one-on-one interviews with participants, building a rapport with them and following up with questions intended to develop a deep understanding of their preferences and opinions. Within the context of an online survey, a virtual moderator emulates the process and principles of a real moderator. They do so by posing dynamic ``probing'' questions to elicit more information related to the ``prime responses'' which participants provide in response to the ``prime questions'' which are preprogrammed in the survey.

In this work, we introduce {\tt SmartProbe}, an API which generates effective probing questions for market research surveys. We describe the modular design of the {\tt SmartProbe} system, including how it leverages the adaptive capacity of large language models (LLMs) to incorporate domain knowledge. Finally, we evaluate the quality of the generated probing questions using expert ratings, and validate the hypothesis that an intelligent virtual moderator can bring tangible benefits to market researchers.

\begin{figure*}[t]
\includegraphics[width=0.99\linewidth]{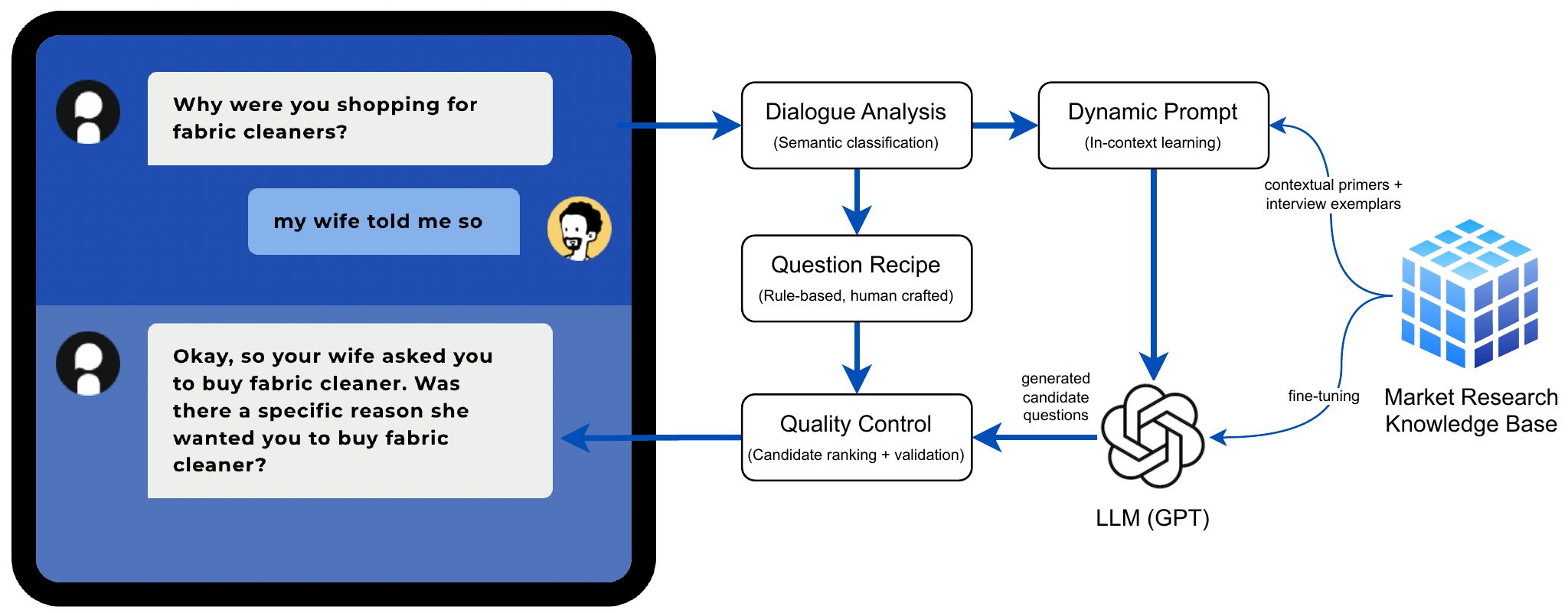}
\caption{Processing flow of our {\tt SmartProbe}, shown with an example input (prime question and response) and output (probing question).
The Dialogue Analysis module first extracts structured textual data from previous conversations between an agent and the user.  Our Dynamic Prompt module and LLM module, incorporating with Market Research Knowledge Base, then generate probing questions. For certain cases when the generated questions are not appropriate, we use the template questions from the Question Recipe module. Finally, the generated question will be checked by the Quality Control module to ensure it is of high quality and meets ethical standards.}
\label{fig:smartprobe}
\end{figure*}

\section{Background and Related Work}

In this section, we briefly review prior work on conversational agents, highlighting the differences between our proposed system and these agents.

In the natural language processing literature, work on conversational agents has tended to focus on task-oriented and chit-chat systems \citep{hussain2019survey, sun2020adding}. Whereas the former are characterized by a limited conversational scope and with predictable user intents and information types, the latter are characterized by a broad conversational scope and with little predictive power in what a user might express. Some efforts have been made to integrate the two together \citep{https://doi.org/10.48550/arxiv.1511.06931, 10.5555/3171837.3171875}. Typically, work on both types of conversational agents tends to focus on generating responses to user utterances, rather than generating questions.

Some literature also focuses on generating questions. For example, considerable advances have been made in generating factoid type questions \citep{serban2016generating}, clarification questions \citep{rao2019answer, zamani2020generating}, and slot-filling questions \citep{patidar2018automatic}. However, relatively little work has focused on generating questions to elicit subjective opinions, and even less work on generating probing questions needed for effective virtual moderation \citep{seltzer2022flipping}.

Most prior works related to conversational agents use sequence-to-sequence (seq2seq) models for data-driven response generation \citep{sordoni-etal-2015-neural, li-etal-2016-diversity}. Recent advances mostly build on transformer-based models \cite{zhang-etal-2020-dialogpt, zhong-etal-2020-towards}.

Experiments with LLMs such as ChatGPT have demonstrated impressive capabilities, including generating good questions: for example, a user could tell ChatGPT to ``\textit{act like an interviewer}'' and give it a few examples of good questions, and it is likely to generate an apt question in a variety of contexts. 
However, in our pilot studies, as with an untrained human tasked with acting like a market researcher, LLMs are likely to make mistakes and ask ineffective or inappropriate questions without proper training and quality control. To address these limitations, in \Cref{sec:API}, we describe the design of our {\tt SmartProbe} API that is able to achieve these goals.

\section{{\tt SmartProbe} API}
\label{sec:API}

In this section, we introduce {\tt SmartProbe}, which can generate probing questions for market research conversations.
As we demonstrate in \Cref{sec:evaluation}, the probing questions generated from our method can lead to deeper insight at scale, since they elicit more useful information from survey participants versus their original answer alone.

To ensure that the probing questions are able to elicit information relevant to the research objectives, {\tt SmartProbe} leverages in-context learning and fine-tuning of GPT to generate candidate probing questions. Generated candidate probing questions are ranked to determine which is the most appropriate probing question; when there is uncertainty about the quality of the generated probing questions, the system falls back to probing questions produced with a rule-based method.

\subsection{Modules}
\label{sec:modules}

{\tt SmartProbe} consists of a set of modules through which the input dialogue is processed, ultimately producing a probing question, with domain expertise provided dynamically through the Market Research Knowledge Base (a set of annotated research conversations), and statically with the usage of Question Recipes (templated probing questions designed by market researchers), as shown in \Cref{fig:smartprobe}.

\paragraph{Dialogue Analysis Module.} 

The research objectives and survey context can be inferred from the dialogue input, or from additional information provided through enhanced context / conversational targets. This analysis module uses information extraction techniques and semantic classification with sentence embeddings, and provides structured contextual data to the Dynamic Prompt Module as well as the Question Recipe Module.

\paragraph{Dynamic Prompt Module.}

Based on prompt engineering experiments across a variety of cues and in-context exemplars, evaluated by market research experts with respect to the quality of the generated probe as well as the richness of the information it elicited from real participants in pilot studies, we have designed effective prompt templates for different research objectives and contexts. This module incorporates the dialogue analysis results as well as the current dialogue, and produces the prompt which serves as input to the LLM Module.

\paragraph{Large Language Model (LLM) Module.}

We employ a model in the InstructGPT family \citep{ouyang2203training}, fine-tuned on exemplars from our Market Research Knowledge Base. 
In our actual usage, the temperature is set to 0 for reproducibility, and a relatively high presence penalty is used to discourage the possibility of generating a verbatim probing question which was used as an in-context exemplar.

\paragraph{Question Recipe Module.}

Sometimes none of the candidate question probes generated with the LLM are entirely appropriate or effective, and badly structured outputs are possible (e.g. generations which do not contain question mark characters). This can be especially true in edge cases, when the dialogue is very concise or extremely detailed. In such cases, relying on a simple but hand-crafted question can be most effective. The Question Recipe Module relies on a bank of prespecified probing question templates, written in each language by a native speaker, which have hand-crafted eligibility requirements and template slot-filling rules which are dynamically processed based on the information extracted from the Dialogue Analysis Module.

\paragraph{Quality Control Module.}

In this module, several simple heuristics are combined to ensure that the generated questions do not contain any toxic language; that they are well-formed questions; and that they are appropriate to the research context. By ranking the candidate probing questions according to these heuristics, this module is responsible for determining which of the generated probing questions and/or question recipes will be most effective for the current dialogue.

\subsection{Features}
\label{sec:features}

\paragraph{Core Features.} 

{\tt SmartProbe} supports more than 14 languages, where for each language we have validated the effectiveness of sentence embeddings and LLM for our analysis and probing question generation purposes, and crafted question recipes written by a native speaker. {\tt SmartProbe} also provides the ability to switch between different personas (e.g., formal versus informal), and to detect and respond appropriately when responses have given a low-effort answer.

\paragraph{Enhanced Context and Target-based Probing.} 

Although {\tt SmartProbe} can usually infer context and objectives based on previous dialogue, it is often beneficial (especially for niche research) for market researchers to provide additional context in the form of research objectives descriptions, conversational targets, and probing question exemplars. Incorporating this information into each of the modules has resulted in higher quality probing questions in some case studies.

\paragraph{Multi-turn Probing.}

It is possible to use {\tt SmartProbe} in a recursive manner, feeding in each new turn of dialogue in order to generate a new probing question. However, we have also seen preliminary results that incorporating a summary of the previous conversation into the LLM prompt can be effective in generating useful probing questions. The market researcher can decide how many turns of probing they want, or this can be determined dynamically. 

\begin{table*}[]
\centering
\resizebox{0.99\textwidth}{!}{
\begin{tabular}{cll}
\toprule
Points & Examples \\\midrule
\multirow{3}{*}{1/5} & Prime Question: Could you tell me why that one   is more likely to pique your interest? \\
  & Prime Response: More simple, and the background match the colour of the brand \\
  & Probing Question: I see. Would you say that the design is more important to you than   the content of the message? \\\midrule
\multirow{3}{*}{2/5} & Prime Question: If you could name ONE reason or   event that prompted your decision to move, what would that be? \\
  & Prime   Response: Highly paid jobs \\
  & Probing   Question: I see. So, you were looking for a better job opportunity? \\\midrule
\multirow{3}{*}{3/5} & Prime Question: Why would you feel more positive   about the brand? \\
  & Prime   Response: I think they would be more environmentally and ethically friendly \\
  & Probing   Question: I see. What makes you say that? \\\midrule
\multirow{3}{*}{4/5} & Prime Question: Can you explain in your own   words, what does your home mean to you? \\
  & Prime   Response: Haven of peace and tranquility! \\
  & Probing   Question: What makes your home a haven of peace and tranquility? \\\midrule
\multirow{3}{*}{5/5} & Prime Question: Why do you buy your food and   drink from Sainsburys? \\
  & Prime   Response: Habits! \\
  & Probing   Question: I see. What is it about Sainsburys that keeps you coming back?\\
 \bottomrule
\end{tabular}
}
\caption{Scoring guideline for evaluating the quality of generated probing questions from {\tt SmartProbe}. Given a pair of prime question and prime response, our {\tt SmartProbe} API will generate a probing question. Our market researchers then mark it using a scale point 1 to 5. Refer to \Cref{tb:scale_points_results} for descriptions of the point system.}
\label{tb:eval_scale_point_def}
\end{table*}

\section{Evaluation}
\label{sec:evaluation}

In order to evaluate the efficacy of {\tt SmartProbe} as a virtual moderator, we wish to understand the extent to which the probing questions that it generates are seen as effective by market researchers; as well as the qualitative benefit of using it as a virtual moderator, with respect to the additional insights that it helps to elicit from survey participants.

\subsection{Quality of {\tt SmartProbe} Generations}
\label{sec:quality}

To validate whether the probing questions generated with {\tt SmartProbe} are considered as effective by market researchers, we evaluated the probing questions in a wide variety of market research contexts commissioned by industry clients including financial services, automotive, consumer packaged goods, public affairs, etc.

\begin{table}[t]
\centering
\resizebox{0.49\textwidth}{!}{
\begin{tabular}{lp{0.46\textwidth}c}
\toprule
Points & Description & Evaluation \\\midrule
1/5 & Irrelevant or leading that might risk biasing the participant's future answers & <1\% \\
2/5 & Simply reiterating the prime answer as an acknowledgment & 18\% \\
3/5 & A generic probing question without referencing the participant's prime answer & 13\% \\
4/5 & A good and sensible probing question referencing the participant's prime answer  & 59\% \\
5/5 & An excellent probing question that really gets the research objectives across  & 10\%\\\bottomrule
\end{tabular}
}
\caption{Percentage breakdowns for the quality of generated probing questions from {\tt SmartProbe}. The percentages are based on 300 probing questions.}
\label{tb:scale_points_results}
\end{table}

\paragraph{Methods.}

For the purpose of this paper, we have included evaluation results from 300 triplets, each consisting of a prime question written by a professional market researcher, a corresponding prime response written by a research participant, and a probing question generated by {\tt SmartProbe}. These triplets cover 50 distinct prime questions across 5 main research categories: Usage and Attitude Research, Advertising Testing, Concept Testing, Customer Experience Research, and Brand Understanding Research. 

We evaluate the quality of the probing questions in the triplets using the 5-point scale listed in \Cref{tb:eval_scale_point_def}. This scoring guideline is developed and refined by three of our in-house market researchers.
In this study, we do not use automatic metrics (such as ROUGE), as the evaluation involves the market research domain knowledge.

\paragraph{Findings.}

Our evaluation results are listed in \Cref{tb:scale_points_results}. The results show that 69\% of probing questions generated by {\tt SmartProbe} are rated as 4/5 or 5/5 by our Market Researchers; while less than 1\% of the probing questions are marked as 1/5. It indicates that {\tt SmartProbe} is usually successful at generating high quality probing questions, and rarely generates very low quality questions

\subsection{Effectiveness of using {\tt SmartProbe} in Quantitative Surveys}
\label{sec:effectiveness}

To evaluate the effectiveness and benefits of using {\tt SmartProbe} in quantitative surveys, we partnered with a leading Canadian research execution company, \textit{The Logit Group}, and conducted a series of self-funded experiments.

We aim to address the following questions that many market researchers may ask: \textbf{(1)} Are probing questions generated by {\tt SmartProbe} better than generic probing questions? \textbf{(2)} What, if any, are the concrete benefits to market researchers when {\tt SmartProbe} is added to quantitative surveys, and how can these benefits help with better decision-making?
In this study, we do not directly evaluate the quality of the generated questions, but rather the responses from these questions, as the ultimate goal of market research surveys is to get effective and informative answers from customers.

\begin{table*}[h!]
\centering
\scriptsize
\resizebox{0.99\textwidth}{!}{
\begin{tabular}{p{0.99\textwidth}}
\toprule
\textbf{\underline{Scale Point 1: Uninformative}} \\
\textit{\underline{Prime Question:}} Could you elaborate on why Going with the flow matters to you the most?  What does it mean to you?   How do you demonstrate that in your life? \\
\textit{\underline{Prime   Response:}} I do not panic or worry. I just take whatever comes by \\
 \textit{\underline{Probing   Question:}} That sounds like a great way to live! Can you give me an example of   a time when you went with the flow? \\
 \textit{\underline{Probing   Response:}} I can't remember right now \\\midrule
 \textbf{\underline{Scale Point 2: Reiterating the First Response/Confirmation}} \\
\textit{\underline{Prime Question:}} Could you elaborate on why Wealth matters to you the most?  What does it mean to you? How do you   demonstrate that in your life? \\
\textit{\underline{Prime Response:}} Just to be independent, for freedom. \\
\textit{\underline{Probing Question:}} I see. What does financial independence mean to you? \\
\textit{\underline{Probing Response:}} Relying only on myself. \\\midrule
\textbf{\underline{Scale Point 3: A Slight Elaboration on the First Response}}\\
\textit{\underline{Prime Question:}} Could you elaborate on why ``Tradition'' matters to you the most?  What does it mean to you? How do you   demonstrate that in your life? \\
\textit{\underline{Prime Response:}} Traditions are important in my life, and it’s something that holds   me to my beliefs \\
\textit{\underline{Probing Question:}} I see. What are some specific traditions that are important to you? \\
\textit{\underline{Probing Response:}} Being faithful and respecting your elders \\\midrule
\textbf{\underline{Scale Point 4: Great, Informative and Illustrative}} \\
\textit{\underline{Prime Question:}} Could you elaborate on why ``Duty'' matters to you the most?  What does it mean to you? How do you   demonstrate that in your life? \\
\textit{\underline{Prime Response:}} I have responsibilities to my family, particularly my wife and our   children. The duties to them will take precedence over almost everything in life. \\
\textit{\underline{Probing Question:}} That is very admirable. Can you give me an example of how you have   put your family first? \\
\textit{\underline{Probing Response:}} There are times when work can get in the way of family time, but I make sure not to bring my work home. I like to be available to my family as   much as I can. \\\midrule
\textbf{\underline{Scale Point 5: Excellent with Rich Stories, Emotions or Examples}}\\
\textit{\underline{Prime Question:}} Could you elaborate on why Authenticity matters to you the most?  What does it mean to you? How do you demonstrate that in your life? \\
\textit{\underline{Prime Response:}} To me, authenticity means being your most organic self. Not changing who you are to fit into a mold society has made. The way I demonstrate authenticity is to always be honest and open with those around me. I don't seek approval from others ... I am who I am. \\
\textit{\underline{Probing Question:}} That's great! Can you give me an example of a time when you were being authentic? \\
\textit{\underline{Probing Response:}} Several years ago my extended family was trying to push me in a certain direction; a direction I didn't want to go in but that they all felt was the right direction for me. While I appreciated that they felt they were making the right choice for me, I kindly let them know that I wasn't interested in pursuing the option they preferred but that I was going to choose my own path instead. \\
\bottomrule
\end{tabular}
}
\caption{Scoring guideline for evaluating the quality of probing responses from real survey participants to the probing questions generated by either {\tt SmartProbe} or pre-programmed generic probing questions. Given a prime question, a prime response and a probing question, our in-house market researchers evaluate the corresponding probing response using a scale point 1 to 5. All provided examples here are from our {\tt SmartProbe}.}
\label{tab:study_1_score}
\end{table*}

\subsubsection{Is {\tt SmartProbe} Better than Generic Probing Questions? }

To address this question, we compare the effectiveness of {\tt SmartProbe} versus the pre-programmed generic probing questions.
The selected generic question has been widely used in current existing market research surveys.  

\paragraph{Methods.}

\textit{The Logit Group} fielded two parallel surveys in January 2023: one used a standard online survey platform, and the other used our INCA conversational survey platform.
Both surveys sampled Canadians and Americans who are nationally representative in terms of age, gender, province/state, and ethnicity. The surveys were conducted in English. The sample sizes were $n=457$ for the standard survey, and $n=500$ for the INCA survey. 

In the standard survey, we asked participants the following series of questions:

1. \textit{Which of the following items do you consider most important to you?} [Close-ended question using rank order: items provided are characteristics such as ``self-growth'' and ``wealth''.]

2. \textit{Why is the item that you ranked first the most important to you?} [Open-ended prime question.]

3. \textit{Why did you say that? Can you elaborate?} [Open-ended probing question.]

In our INCA survey, we asked the same questions, except that the third question was generated based on the participants' previous answers using {\tt SmartProbe}.

We then evaluated the quality of participants' responses to the probing questions in each of the surveys using the 5-point scale defined in \Cref{tab:study_1_score}.

\paragraph{Findings.}

Our results are listed in \Cref{tb:comparsion}. We observe that {\tt SmartProbe} is much more likely than generic probing questions to elicit high-quality responses. From our evaluation, 76\% of the responses elicited by {\tt SmartProbe} were annotated as either 4/5 or 5/5, versus 25\% of responses when using generic probing. These findings demonstrate that {\tt SmartProbe} is able to elicit additional information conducive to deeper understanding.

\begin{table}[]
\centering
\resizebox{0.48\textwidth}{!}{
\begin{tabular}{lcc}
\toprule
Scale Point & Standard & INCA \\\midrule
1 - Uninformative & 14\%$^\dagger$ & 2\%\\
2 - Reiterating the first response/confirmation & 39\%$^\dagger$ & 15\% \\
3 - A slight elaboration on the first response & 22\%$^\dagger$ & 7\% \\
4 - Great, informative, and illustrative & 14\% & 26\%$^\dagger$ \\
5 - Excellent with rich stories, emotions or examples & 11\% & 50\%$^\dagger$ \\
\bottomrule
\end{tabular}
}
\caption{Percentage breakdowns for the quality of probing responses to the probing questions generated either by {\tt SmartProbe} (in the INCA survey) or pre-programmed generic probing questions (in the standard survey). The sample sizes are $n=457$ for the standard survey and $n=500$ for the INCA survey.  $^\dagger$Significantly higher with two Sample $Z$ test at 95\% confidence level.}
\label{tb:comparsion}
\end{table}

\subsubsection{What are Concrete Benefits When Using {\tt SmartProbe}?}

\paragraph{Methods.}

\textit{The Logit Group} also fielded an INCA survey as part of a large study in January 2023, with a nationally representative sample of $n=1,231$ Canadians. 
The survey was conducted in English and French. We use the following three questions for the analyses: 

1. ``\textit{What is the first thing that comes to mind, when you think about Justin Trudeau/Pierre Poilievre/Jagmeet Singh?}'' [Open-ended prime question.]

2. [Probing question generated by {\tt SmartProbe} based on the prime question and response.]

3. ``\textit{If a federal election were held today, you would vote for ... ?}'' [Close-ended question with the three leaders and their parties shown as options for participants to choose from.]

We then annotated the prime responses as well as the combined responses into themes, using a shared theme list and methodology. The annotation was conducted using a semi-automated approach which generated clusters based on sentence embeddings, which was then adjusted into a theme list and corrected by market researchers.

We analyzed the themes across the 3 political leaders and conducted a significance test to compare the leaders. We performed this analysis for both prime responses and combined responses.

Finally, with respect to the close-ended voting intention question, we ran a driver analysis to see which open-ended themes drive voter intention. Again we performed this analysis for both prime responses and combined responses.

\paragraph{Findings.}

By comparing the results between the prime responses and the combined responses, we observe three benefits that {\tt SmartProbe} provides: 

(1) {\tt SmartProbe} leads to more considered answers: 
We observe higher average word count in combined responses (with probing) than in prime responses (without probing).  We also found less uninformative answers such us ``don't know / not sure'' or ``reference a political party/leader without any detail/sentiment'' in combined responses than prime responses.

(2)  {\tt SmartProbe} led to more differentiated findings:
We found twice as many significant differences between the 3 political leaders in combined responses than in prime responses alone. The responses to the probing questions were more specific and much more likely to be focused on specific policies which differ between the leaders. 

(3) {\tt SmartProbe} led to actionable insights:
A much greater number of strengths and improvement areas related to voting intention were identified from the themes elicited from the combined responses than the prime responses. Such insights provide concrete directions for the leaders in order to be effective in swaying voting intention.  

\section{Conclusion}

In this work, we propose adding {\tt SmartProbe} to quantitative surveys to fill the role of a virtual moderator as an effective means of bridging the gap between qualitative and quantitative market research. The modular processing flow of {\tt SmartProbe}, designed and developed with extensive market research domain expertise, proves to be successful in generating high-quality probing questions that elicit more considered responses, more differentiated findings and more actionable results. The capacity for {\tt SmartProbe} to elicit rich information at scale may ultimately lead to better and more confident decision making from online surveys.

\section*{Ethical Considerations}

As our probing question is generated based on LLM model and directly interacts with users, we design a quality control module to ensure the generated questions from our API follow ethical standards (discussed in \Cref{sec:modules}).
All our real-world evaluations for our {\tt SmartProbe} are conducted in partnership with \textit{The Logit Group}, a leading Canadian research execution company.

\bibliography{ref}

\begin{thebibliography}{14}
\expandafter\ifx\csname natexlab\endcsname\relax\def\natexlab#1{#1}\fi

\bibitem[{Dodge et~al.(2015)Dodge, Gane, Zhang, Bordes, Chopra, Miller, Szlam,
  and Weston}]{https://doi.org/10.48550/arxiv.1511.06931}
Jesse Dodge, Andreea Gane, Xiang Zhang, Antoine Bordes, Sumit Chopra, Alexander
  Miller, Arthur Szlam, and Jason Weston. 2015.
\newblock Evaluating prerequisite qualities for learning end-to-end dialog
  systems.
\newblock \emph{arXiv preprint arXiv.1511.06931}.

\bibitem[{Hussain et~al.(2019)Hussain, Ameri~Sianaki, and
  Ababneh}]{hussain2019survey}
Shafquat Hussain, Omid Ameri~Sianaki, and Nedal Ababneh. 2019.
\newblock A survey on conversational agents/chatbots classification and design
  techniques.
\newblock In \emph{Web, Artificial Intelligence and Network Applications:
  Proceedings of the Workshops of the 33rd International Conference on Advanced
  Information Networking and Applications (WAINA-2019) 33}, pages 946--956.
  Springer.

\bibitem[{Li et~al.(2016)Li, Galley, Brockett, Gao, and
  Dolan}]{li-etal-2016-diversity}
Jiwei Li, Michel Galley, Chris Brockett, Jianfeng Gao, and Bill Dolan. 2016.
\newblock A diversity-promoting objective function for neural conversation
  models.
\newblock In \emph{Proceedings of the 2016 Conference of the North {A}merican
  Chapter of the Association for Computational Linguistics: Human Language
  Technologies}, pages 110--119, San Diego, California. Association for
  Computational Linguistics.

\bibitem[{Ouyang et~al.(2022)Ouyang, Wu, Jiang, Almeida, Wainwright, Mishkin,
  Zhang, Agarwal, Slama, Ray et~al.}]{ouyang2203training}
Long Ouyang, Jeff Wu, Xu~Jiang, Diogo Almeida, Carroll~L Wainwright, Pamela
  Mishkin, Chong Zhang, Sandhini Agarwal, Katarina Slama, Alex Ray, et~al.
  2022.
\newblock Training language models to follow instructions with human feedback.
\newblock \emph{arXiv preprint arXiv:2203.02155}.

\bibitem[{Patidar et~al.(2018)Patidar, Agarwal, Vig, and
  Shroff}]{patidar2018automatic}
Mayur Patidar, Puneet Agarwal, Lovekesh Vig, and Gautam Shroff. 2018.
\newblock Automatic conversational helpdesk solution using seq2seq and
  slot-filling models.
\newblock In \emph{Proceedings of the 27th ACM International Conference on
  Information and Knowledge Management}, pages 1967--1975.

\bibitem[{Rao and Daum{\'e}~III(2019)}]{rao2019answer}
Sudha Rao and Hal Daum{\'e}~III. 2019.
\newblock Answer-based adversarial training for generating clarification
  questions.
\newblock \emph{arXiv preprint arXiv:1904.02281}.

\bibitem[{Seltzer et~al.(2022)Seltzer, Cheng, Zong, and
  Lin}]{seltzer2022flipping}
Josh Seltzer, Kathy Cheng, Shi Zong, and Jimmy Lin. 2022.
\newblock Flipping the script: Inverse information seeking dialogues for market
  research.
\newblock In \emph{Proceedings of the 45th International ACM SIGIR Conference
  on Research and Development in Information Retrieval}, pages 3380--3383.

\bibitem[{Serban et~al.(2016)Serban, Garc{\'\i}a-Dur{\'a}n, Gulcehre, Ahn,
  Chandar, Courville, and Bengio}]{serban2016generating}
Iulian~Vlad Serban, Alberto Garc{\'\i}a-Dur{\'a}n, Caglar Gulcehre, Sungjin
  Ahn, Sarath Chandar, Aaron Courville, and Yoshua Bengio. 2016.
\newblock Generating factoid questions with recurrent neural networks: The 30m
  factoid question-answer corpus.
\newblock \emph{arXiv preprint arXiv:1603.06807}.

\bibitem[{Sordoni et~al.(2015)Sordoni, Galley, Auli, Brockett, Ji, Mitchell,
  Nie, Gao, and Dolan}]{sordoni-etal-2015-neural}
Alessandro Sordoni, Michel Galley, Michael Auli, Chris Brockett, Yangfeng Ji,
  Margaret Mitchell, Jian-Yun Nie, Jianfeng Gao, and Bill Dolan. 2015.
\newblock A neural network approach to context-sensitive generation of
  conversational responses.
\newblock In \emph{Proceedings of the 2015 Conference of the North {A}merican
  Chapter of the Association for Computational Linguistics: Human Language
  Technologies}, pages 196--205, Denver, Colorado. Association for
  Computational Linguistics.

\bibitem[{Sun et~al.(2020)Sun, Moon, Crook, Roller, Silvert, Liu, Wang, Liu,
  Cho, and Cardie}]{sun2020adding}
Kai Sun, Seungwhan Moon, Paul Crook, Stephen Roller, Becka Silvert, Bing Liu,
  Zhiguang Wang, Honglei Liu, Eunjoon Cho, and Claire Cardie. 2020.
\newblock Adding chit-chat to enhance task-oriented dialogues.
\newblock \emph{arXiv preprint arXiv:2010.12757}.

\bibitem[{Yu et~al.(2017)Yu, Black, and Rudnicky}]{10.5555/3171837.3171875}
Zhou Yu, Alan~W. Black, and Alexander~I. Rudnicky. 2017.
\newblock Learning conversational systems that interleave task and non-task
  content.
\newblock In \emph{Proceedings of the 26th International Joint Conference on
  Artificial Intelligence}, IJCAI'17, page 4214–4220. AAAI Press.

\bibitem[{Zamani et~al.(2020)Zamani, Dumais, Craswell, Bennett, and
  Lueck}]{zamani2020generating}
Hamed Zamani, Susan Dumais, Nick Craswell, Paul Bennett, and Gord Lueck. 2020.
\newblock Generating clarifying questions for information retrieval.
\newblock In \emph{Proceedings of the Web Conference 2020}, pages 418--428.

\bibitem[{Zhang et~al.(2020)Zhang, Sun, Galley, Chen, Brockett, Gao, Gao, Liu,
  and Dolan}]{zhang-etal-2020-dialogpt}
Yizhe Zhang, Siqi Sun, Michel Galley, Yen-Chun Chen, Chris Brockett, Xiang Gao,
  Jianfeng Gao, Jingjing Liu, and Bill Dolan. 2020.
\newblock {DIALOGPT} : Large-scale generative pre-training for conversational
  response generation.
\newblock In \emph{Proceedings of the 58th Annual Meeting of the Association
  for Computational Linguistics: System Demonstrations}, pages 270--278,
  Online. Association for Computational Linguistics.

\bibitem[{Zhong et~al.(2020)Zhong, Zhang, Wang, Liu, and
  Miao}]{zhong-etal-2020-towards}
Peixiang Zhong, Chen Zhang, Hao Wang, Yong Liu, and Chunyan Miao. 2020.
\newblock Towards persona-based empathetic conversational models.
\newblock In \emph{Proceedings of the 2020 Conference on Empirical Methods in
  Natural Language Processing (EMNLP)}, pages 6556--6566, Online. Association
  for Computational Linguistics.

\end{thebibliography}
\bibliographystyle{acl_natbib}




\end{document}